\documentclass[letterpaper]{article} % DO NOT CHANGE THIS
\usepackage{aaai24}  % DO NOT CHANGE THIS
\usepackage{times}  % DO NOT CHANGE THIS
\usepackage{helvet}  % DO NOT CHANGE THIS
\usepackage{courier}  % DO NOT CHANGE THIS
\usepackage[hyphens]{url}  % DO NOT CHANGE THIS
\usepackage{graphicx} % DO NOT CHANGE THIS
\usepackage{natbib}  % DO NOT CHANGE THIS AND DO NOT ADD ANY OPTIONS TO IT
\usepackage{caption} % DO NOT CHANGE THIS AND DO NOT ADD ANY OPTIONS TO IT
\usepackage{amssymb}
\usepackage{amsthm}
\usepackage{amsmath}
\usepackage{algorithm}
\usepackage{algorithmic}
\usepackage{color}
\usepackage{bm}
\usepackage{subfigure}
\usepackage{newfloat}
\usepackage{listings}
\DeclareCaptionStyle{ruled}{labelfont=normalfont,labelsep=colon,strut=off} % DO NOT CHANGE THIS
\floatstyle{ruled}
\newfloat{listing}{tb}{lst}{}
\floatname{listing}{Listing}
\title{Live and Learn: Continual Action Clustering with Incremental Views}
\author{
    Xiaoqiang Yan\textsuperscript{\rm 1},
    Yingtao Gan\textsuperscript{\rm 1},
    Yiqiao Mao\textsuperscript{\rm 1},
    Yangdong Ye\textsuperscript{\rm 1}\thanks{Corresponding author},
    Hui Yu\textsuperscript{\rm 2}
}
\affiliations{
    \textsuperscript{\rm 1} School of Computer and Artificial Intelligence, Zhengzhou University, Zhengzhou, China\\
    \textsuperscript{\rm 2} School of Creative Technologies, University of Portsmouth, PO1 2DJ, United Kingdom\\
     iexqyan@zzu.edu.cn, ieytgan@outlook.com, ieyqmao@gs.zzu.edu.cn, ieydye@zzu.edu.cn, hui.yu@port.ac.uk
}

\usepackage{bibentry}
\begin{document}

\maketitle

\begin{abstract}
  Multi-view action clustering leverages the complementary information from different camera views to enhance the clustering performance. Although existing approaches have achieved significant progress, they assume all camera views are available in advance, which is impractical when the camera view is incremental over time. Besides, learning the invariant information among multiple camera views is still a challenging issue, especially in continual learning scenario. Aiming at these problems, we propose a novel  continual action clustering (CAC) method, which is capable of learning action categories in a continual learning manner. To be specific, we first devise a category memory library, which captures and stores the learned categories from historical views. Then, as a new camera view arrives, we only need to maintain a consensus partition matrix, which can be updated by leveraging the incoming new camera view rather than keeping all of them. Finally, a three-step alternate optimization is proposed, in which the category memory library and consensus partition matrix are  optimized. %Besides, the convergence of the proposed CAC.
  The empirical experimental results on 6 realistic multi-view action collections demonstrate the excellent clustering performance and time/space efficiency of the CAC compared with 15 state-of-the-art baselines.
\end{abstract}

\section{Introduction}
Recognizing human actions from videos is a fundamental and active research topic in various applications. %such as video summarization and event retrieval~\cite{DBLP:journals/tip/LiuMHVRK22}.
In recent years, action recognition has achieved significant attentions and promising progresses due to the explosive growth of action data and the advancement of relevant learning models.
Usually, most existing methods focus on supervised learning scenarios which leverage sufficient data labels to train learning models. However, labelling large amounts of data is time-consuming and labor-intensive. In addition, due to the different backgrounds and prior knowledge of annotators, they cannot assure to uniformly assign action labels to similar video samples. Thus, it is necessary to resort to unsupervised clustering approaches to automatically learn action categories.

In last decades, numerous action clustering approaches have been devised and achieved some satisfactory performances. For instance, \cite{DBLP:conf/cvpr/JonesS14} designs a dual assignment $k$-means to perform two co-occurring action clustering tasks. \cite{DBLP:conf/ijcai/BhatnagarSAJ17} proposes a generic unsupervised approach to learn feature representation for first person action clustering. \cite{yan2017multi} proposes to perform action clustering in a multi-task clustering setting. \cite{DBLP:journals/tii/PengLFSH20} proposes to improve the performance of action clustering with contextual information of actions in a recursive constrained framework. \cite{DBLP:conf/cvpr/KumarHAKZT22} performs representation learning and online action clustering for unsupervised activity segmentation. However, existing action clustering faces the following issues: 1) Difficulty in action category discovery. Most current approaches focus on making better use of spatio-temporal information to describe the actions. However, discovering action categories remains a challenging problem due to camera motions, changes of viewpoints, cluttered background. 2) Ignoring the correlations between multiple camera views. Due to the occlusion and deformation of the viewpoints, single-view action clustering cannot effectively capture the complementary information between multiple views. For example, it is impossible to recognize applauding solely from the camera behind a person. Thus, it is natural to perform multi-view clustering on the task of recognizing human actions.

Multi-view clustering (MVC) leverages the complementary information from different camera views to enhance the performance of action clustering~\cite{DBLP:journals/tmm/XiaWGZG22}. According to the strategy of complementary information extraction, existing MVC approaches can be classified into the following types: subspace clustering~\cite{8387526,DBLP:conf/aaai/MaoYGY21,TANG2023333,Yan_2023_CVPR,DBLP:journals/tip/XuLPRSSZ23,DBLP:journals/isci/ZhaoYN23,10130781}, consensus clustering~\cite{DBLP:conf/ijcai/WangLZTLHXY19,DBLP:conf/icml/Liu0LWZTT0Z21,DBLP:journals/corr/abs-2203-11572,10128150} and multi-kernel clustering~\cite{DBLP:conf/aaai/LiuDY0Z16,DBLP:conf/aaai/GuoY19,DBLP:journals/tmm/ZhangLGWNS23,DBLP:journals/pami/Liu23}. Although existing MVC approaches have achieved some promising results, they assume that all camera views are available in advance, which is impractical when the camera view is incremental over time. Besides, learning the invariant information among multiple camera views is a difficult and challenging issue in continual learning scenario due to the following reasons. First, there is a large divergence between historical and new coming views usually. Second, the new coming views are often unknown and unpredictable to historical views. Third, it is impractical to store all historical views for directly employing existing MVC approaches to re-access all views simultaneously.

\begin{figure}[t]
  \begin{center}
     \includegraphics[width=0.8\linewidth]{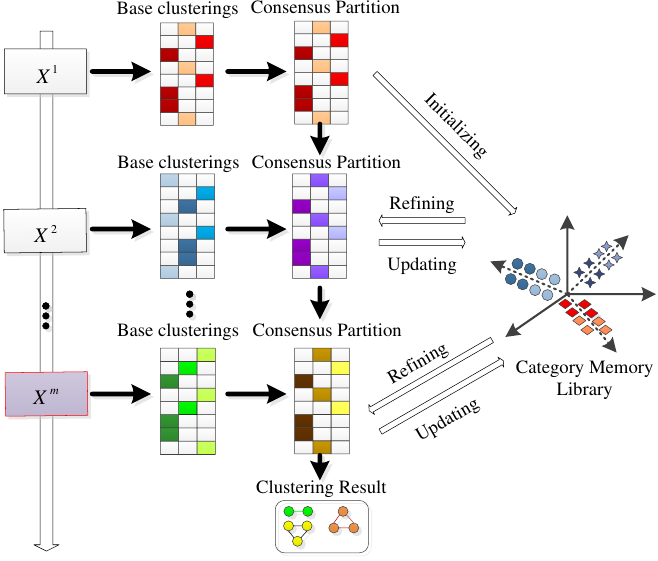}
  \end{center}
  \caption{Framework of CAC. When a new view arrives, we only need to maintain a consensus partition matrix to achieve never-ending knowledge transfer between historical and new coming views with the category memory library.}\label{model}
  \vspace{-0.3cm}
\end{figure}

To solve these problems, we propose a novel continual action clustering (CAC) method, which is capable of achieving never-ending knowledge transfer between historical views and the new coming ones (as shown in Figure~\ref{model}). To achieve this target, we first design a category memory library, which learns and stores the learned action categories from historical views. When a new camera view arrives, CAC constructs its basic clustering partitions based on kernel learning. Then, we only need to maintain a consensus partition matrix for historical views, which can be updated by leveraging the incoming new camera view rather than keeping all of them. Finally, a three-step alternate optimization is proposed, in which the category memory library and consensus partition matrix are alternately optimized. Extensive experimental results demonstrate the clustering performance and time/space efficiency of the proposed CAC. The contributions of this study can be summarized as follows:
\begin{itemize}
  \item A novel continual action clustering method named CAC is proposed, which can achieve never-ending knowledge transfer between historical views and the new coming ones, and improve the clustering performance accordingly. To the best of our knowledge, this is the first work to explore the task of multi-view action clustering in incremental learning scenario.
  \item We design a category memory library to learn and store the learned action categories from historical views. When new action view is coming, we only need to maintain a consensus partition matrix for historical views, which can be updated by leveraging the incoming new camera view rather than keeping all of them.
  \item A three-step alternate optimization is proposed, in which the category memory library and consensus partition matrix are alternately optimized. The experimental results strongly support the proposed CAC method.
\end{itemize}

\section{Related Works and Backgrounds}
%\subsection{Action Clustering}
\subsection{Incremental Multi-view Clustering}

%Although existing MVC approaches have been studied extensively in the aforementioned aspects, they assume that all view are available in advance, which is impractical when the view is incremental in time.
In recent several years, there are few efforts towards solving incremental MVC so far~\cite{DBLP:journals/kbs/ZhouSDYL19,DBLP:journals/nn/YinHZLM21,DBLP:journals/pami/SunCDLDY22,DBLP:conf/mm/WanLLLWZ22}. Specifically, \cite{DBLP:journals/kbs/ZhouSDYL19} divides all views into several incremental subspaces and then performs $k$-means iteratively by increasing the size of the subspaces. \cite{DBLP:journals/nn/YinHZLM21} extends spectral clustering to incremental multi-view clustering setting, which finds a ensemble kernel from existing basic kernels and a new coming view. \cite{DBLP:conf/mm/WanLLLWZ22} first constructs a consensus similarity matrix of all available views, and reconstructs the consensus similarity matrix when newly views are available. Despite the success of existing incremental MVC, they seem far from fully being explored and can still be improved from the following considerations. First of all, with limited computational resources, it would be difficult to access and re-compute all samples in historical views. Secondly, the new coming views are often unknown and unpredictable to historical views, so it is challenging to capture the large divergence between them.

%1) In respect to ``what to learn", existing incremental MVC emphasizes the consistent information of multiple views and ignores the redundant information in each view. To learn multi-view consistency and avoid the misleading of meaningless redundant information, a compact and discriminative representation without redundancy is required, especial for measuring the consistency between old and new views. 2) In respect to ``when to learn", existing approaches assume that the available and new views are related and lack mechanism to discover the views with negative correlations. 3) In respect to ``how to learn", existing methods perform incremental MVC with statistical clustering and separate cluster assignments from feature learning, which results in the trained feature representations not being friendly to the later clustering task.

%To continually learn a new clustering task based on previous experience or knowledge, several life-long clustering approaches have been developed, such as life-long spectral clustering, life-long visual-tactile clustering, life-long matrix factorization. However, these existing life-long clustering cannot directly deal with the task of multi-view clustering. In recent several years, there are few efforts towards solving incremental MVC so far, which have achieved promising results on various applications [][][].

\subsection{Multi-kernel $k$-means}
Traditional $k$-means  cannot effectively deal with the feature transformation of nonlinear data. Thus, kernel $k$-means~\cite{DBLP:journals/pami/Liu23} has been proposed to map nonlinear data into a high dimensional space, in which the data can be separated. Kernel $k$-means has also been applied to the task of multi-view clustering and obtained promising performance, i.e., multi-kernel $k$-means~\cite{DBLP:conf/ijcai/WangLZTLHXY19,DBLP:conf/icml/Liu0LWZTT0Z21}.

Let $\textbf{X}=\{X^i\}_{i=1}^m$ be a data collection with $m$ views, each view contains $n$ data samples $\{\textbf{x}_j\}_{j=1}^n$, where $n$ is the number of data samples. We use $\phi_p(\cdot):\textbf{x}\in \mathcal{X} \rightarrow \mathcal{H}_p$ to denote the mapping function that transforms the data samples in the $p$-th view into reproducing kernel Hilbert space $\mathcal{H}_p(1\leq p \leq m)$. In multi-kernel condition, the data samples of the $m$ views can be represented as $\phi_{\beta}(\textbf{x})=[\beta_1\phi_1(\textbf{x})^T,\cdots,\beta_m\phi_m(\textbf{x})^T]^T$, where $\bm{\beta} =[\beta_1,\cdots,\beta_m]^T$ is the coefficients of $m$ basic kernels $\{\mathcal{K}_p(\cdot,\cdot)\}_{p=1}^m$. According to the definition of $\phi_{\bm{\beta}}(\textbf{x})$, the kernel function $\mathcal{K}_{\bm{\beta}}(\textbf{x}_i,\textbf{x}_j)\}$ can be defined as follows,
\begin{equation}
\begin{aligned}
& \mathcal{K}_{\beta}(\textbf{x}_i,\textbf{x}_j)  = \phi_{\beta}(\textbf{x}_i)^T\phi_{\beta}(\textbf{x}_j) = \sum_{p=1}^m \beta_p^2\mathcal{K}_p(\textbf{x}_i,\textbf{x}_j)\\
\end{aligned}
\end{equation}
where a kernel matrix $\textbf{K}_{\bm{\beta}}$ can be obtained by applying the kernel function on the data collection $\textbf{X}=\{X^i\}_{i=1}^m$.

Multi-kernel $k$-means supposes that the optimal kernel matrix can be characterized by a linear combination of multiple kernel matrices, i.e., $\textbf{K}_{\bm{\beta}}= \sum_{p=1}^m \beta_p^2 \textbf{K}_p$. Thus, we can obtain the objective function of multi-kernel $k$-means,
\begin{equation}\label{}
\begin{aligned}
 &\min_{\textbf{H},\bm{\beta}} {\rm Tr}\left( \textbf{K}_{\bm{\beta}}(\textbf{I}_n - \textbf{H}\textbf{H}^T) \right),\\
 &{\rm s.t.} \textbf{H}\in\mathbb{R}^{n\times k}, \textbf{H}^T\textbf{H} = \textbf{I}_k, \bm{\beta}^T\textbf{1}_m=1, \bm{\beta}_p\geq 0, \forall p. \\
\end{aligned}
\end{equation}
where $\textbf{H}$ is the clustering partition matrix, which can be obtained by alternately optimizing $\bm{\beta}$ and $\textbf{H}$.

\subsection{Late Fusion Multi-view Clustering}
Recently, late fusion MVC methods have been proposed to address the issues of high complexity and cumbersome optimization process of multi-kernel $k$-means. Differently, late fusion MVC separately performs kernel $k$-means or spectral clustering on the kernel matrix $\{\textbf{K}\}_{p=1}^m$ to generate basic clustering partitions $\textbf{H} = \{{\rm H}_p\}_{p=1}^m$ for each view. Then, it searches an optimal partition matrix as follows,
\begin{equation}\label{late}
\begin{split}
 &\max_{\textbf{H},\bm{\beta}} {\rm Tr}({\textbf{H}^*}^T\textbf{X}) + \lambda {\rm Tr}({\textbf{H}^*}^T\textbf{M}),   \\
 &{\rm s.t.} {\textbf{H}^*}^T\textbf{H}^* =  \textbf{I}_k, \textbf{W}_p^T\textbf{W}_p=  \textbf{I}_k,\\
 & \sum_{p=1}^m \beta_p^2 =1, \beta_p\geq 0, \textbf{X} = \sum_{p=1}^m \beta_p \textbf{H}_p\textbf{W}_p.
\end{split}
\end{equation}
where $\beta_p$ is the weight of the $p$-th view, $\textbf{W}_p$ is the permutation matrix of the $p$-th view which can align the basic partition matrices of different views. $\textbf{M}$ is the averaged partition matrix, $\textbf{X} = \sum_{p=1}^m \beta_p \textbf{H}_p\textbf{W}_p$ is the linear combination of the partition matrices from different views. ${\rm Tr}({\textbf{H}^*}^T\textbf{M})$ is a regularization term that prevents the optimal partition matrix from being too far away from the average partition matrix $\textbf{M}$.

\section{Continual Action Clustering with Incremental Views}
In this section, we first devise a category memory library to learn and store the learned action categories from historical views. Then, the objective function of CAC is provided, which is solved by a three-step alternate optimization. Finally, the convergence and complexity is analyzed.

\subsection{Problem Statement}
Given a data collection with $m$ views $\textbf{X}=\{X^i\}_{i=1}^m$, each view contains $n$ data samples $\{\textbf{x}_j\}_{j=1}^n$. The original intention of late fusion multi-view clustering based on multi-kernel $k$-means is to uncover the correlations among all views and generate the final partition matrix $\textbf{H}^*$ by combining the basic partition matrices $\textbf{H} = \{{\rm H}_p\}_{p=1}^m$ of each view. However, this learning strategy may be impractical in the incremental scenario of multi-view action clustering, in which the camera view is incremental over time. In this study, we propose a continual action clustering (CAC) method with incremental views. In incremental multi-view clustering, the CAC encounters a series of views $X^1,X^2,\cdots,X^m$.
When it receives a view of data for clustering in each period, this system should meet the following requirements. 1) \textbf{Continual clustering}. It should achieve never-ending knowledge transfer between historical views and the new coming ones. 2) \textbf{Clustering performance}. The final clustering partition matrix $\textbf{H}^*$ by continually combining the basic ones $\textbf{H} = \{{\rm H}_p\}_{p=1}^m$  should be better that the clustering quality of traditional multi-view clustering. 3) \textbf{Computational speed}. New basic clustering assignment should be arbitrarily and efficiently added when the continual clustering system faces with new views.

\subsection{Category Memory Library}
To conduct the action clustering with incremental views, an important issue is to learn and store the categories (or clustering assignments) from historical views. The objective function Eq.~\ref{late} is a widely used formulation for late fusion multi-view clustering. Except for its simplicity and efficiency, it can also maximize the alignment of the optimal partition matrix by linearly combining the basic partition matrices for each view. However, the linear combination  fails to integrate the basic partition matrices one by one and cannot reveal a more intrinsic clustering structure during the optimization process, which means that it cannot be directly applied to incremental multi-view clustering.

To solve this issue, we devise a category memory library to learn and store the latent cluster centroids  of the basic partition matrix learned from the historical views. %When a new view arrived, we only need to maintain a consensus partition matrix, which can be updated by leveraging the incoming new camera view rather than keeping all of them.
Specifically, we decompose the optimal partition matrix in Eq.~\ref{late} into two sub matrices as follows,
\begin{equation}\label{memory}
\begin{split}
 &\textbf{H}^* = \textbf{E}\textbf{B},
\end{split}
\end{equation}
where $\textbf{E}\in\{0,1\}^{n\times k}$ indicates which category the data belongs to, $\textbf{B}\in \mathbb{R}^{k\times k}$ is our category memory library that stores the cluster centroids in optimal clustering partition matrix. By storing and optimizing the category memory library, it can serve continual action clustering. According to Eq.~\ref{late} and Eq.~\ref{memory}, we can formulate the objective function of late fusion multi-view clustering with a category memory library as follows,
\begin{equation}\label{memory1}
\begin{split}
 &\max_{\textbf{E}, \textbf{B},\{\textbf{W}_p\}_{p=1}^m,\bm{\beta}}  {\rm Tr}(\textbf{B}^T\textbf{E}^T\textbf{X}) + \lambda {\rm Tr}(\textbf{B}^T\textbf{E}^T\textbf{M}),   \\
 &{\rm s.t.} \textbf{B}^T\textbf{B} =  \textbf{I}_k, \textbf{W}_p^T\textbf{W}_p=  \textbf{I}_k, \textbf{E}\in\{0,1\}^{n\times k},\\
 &  \sum_{p=1}^m \beta_p^2 =1, \beta_p\leq 0, \textbf{X} = \sum_{p=1}^m \beta_p \textbf{H}_p\textbf{W}_p.
\end{split}
\end{equation}

The orthogonal constraints imposed on $\textbf{B}$ make each cluster center of $\textbf{B}$ independent and orthogonal to each other, facilitating subsequent optimization. Thus, the category memory library $\textbf{B}$ can be used to store more refined clustering structures, achieving better clustering results while preserving learned knowledge from historical views.

\subsection{Continual Action Clustering}
Although Eq.~\ref{memory1} can store the learned action categories, it still requires to fuse the basic clustering partition matrices of all views at once without considering the significant consumption of space and time in the case of incremental multi-view clustering.
As discussed, existing multi-view action clustering methods improve the clustering performance by fusing information from all views. However, such methods cannot handle the situations where the camera views continue to increase over time. For example, we can obtain new action data when a new camera is installed. Traditional methods require to fuse all views in one time, which is impractical since it wastes a lot of time and space resources.

To address this issue, the proposed CAC creatively combines multi-view action clustering with continual learning. Inspired by the  multi-view clustering approaches based on the late fusion strategy, the proposed CAC only needs to maintain a consensus partition matrix, which can be updated by leveraging the incoming new camera view rather than keeping all of them.

Suppose we have collected the action data of the $t$-th view camera, the basic clustering partition matrix $\textbf{H}_t$ can be generated by applying kernel $k$-means on the $t$-th view. At this time, suppose we already have the consensus partition matrix $\textbf{H}_{t-1}^*$ of the previous $t-1$ cameras. To improve the efficiency of incremental MVC, we just use the optimal permutation matrix $\textbf{W}_t$ of the basic partition matrix $\textbf{H}_t$ to align the partition matrices. Besides, we treat the consensus partition matrix $\textbf{H}_{t-1}^*$ of the previous $t-1$ view cameras as the regularization term rather than the averaged partition matrix $\textbf{M}$. Thus, the  optimal permutation matrix $\textbf{W}_t$ can be obtained  by learning the basic partition $\textbf{H}_t$ after a linear transformation $\textbf{H}_t\textbf{W}_t$ as follows,
\begin{equation}\label{permutation}
\begin{split}
 &\max_{\textbf{E}, \textbf{B}, \textbf{W}_t}  {\rm Tr}(\textbf{B}^T\textbf{E}^T \textbf{H}_t\textbf{W}_t) + \lambda {\rm Tr}(\textbf{B}^T\textbf{E}^T\textbf{H}_{t-1}^*),   \\
 &{\rm s.t.} \textbf{B}^T\textbf{B} =  \textbf{I}_k, \textbf{W}_t^T\textbf{W}_t=  \textbf{I}_k, \textbf{E}\in\{0,1\}^{n\times k}
\end{split}
\end{equation}

As shown in Eq.~\ref{permutation}, the correlation between linear transformation $\textbf{H}_t\textbf{W}_t$ and the consensus partition matrix $\textbf{H}_{t-1}^*$ can be further refined by the category memory library $\textbf{B}$. After obtaining the optimal permutation matrix $\textbf{W}_t$, CAC method can combine  $\textbf{H}_t\textbf{W}_t$ and $\textbf{H}_{t-1}^*$ as follows,
\begin{equation}\label{combine}
%\begin{split}
 \textbf{H}_{t}^* = \textbf{H}_{t-1}^* + \textbf{H}_t\textbf{W}_t
%\end{split}
\end{equation}

By normalizing $\tilde{\textbf{H}}^*_{t-1} = \textbf{H}_{t}^*/(t-1)$, the final version of the CAC objective function can be formulated as follows,
\begin{equation}\label{OBJ}
\begin{split}
 &\max_{\textbf{E}, \textbf{B}, \textbf{W}_p}  {\rm Tr}(\textbf{B}^T\textbf{E}^T\textbf{H}_t\textbf{W}_t) + \lambda {\rm Tr}(\textbf{B}^T\textbf{E}^T\tilde{\textbf{H}}_{t-1}^*),   \\
 &{\rm s.t.} \textbf{B}^T\textbf{B} =  \textbf{I}_k, \textbf{W}_t^T\textbf{W}_t=  \textbf{I}_k, \textbf{E}\in\{0,1\}^{n\times k}
\end{split}
\end{equation}

Now, the proposed CAC can deal with the views one by one with a category memory library.

\begin{algorithm}[t!]
   \caption{Continual Action Clustering (CAC) Algorithm}
   \label{CAC}
\begin{algorithmic}[1]
   \STATE{\bfseries Input:} The basic partition matrices $\{\textbf{H}_t\}_{t=1}^m$ tasks with $T$ datasets $X=\{X^1, X^2, \cdots, X^T\}$, categories number $k$, regularization parameter $\lambda$, convergence boundary $\varepsilon$
\STATE \textbf{Initialization:} Initialize the category memory library $\textbf{B}$ by performing $k$-means on $\textbf{H}_1$
\STATE Initialize the consensus partition matrix by $\textbf{H}_1^* = \textbf{H}_1$
\FOR{$t=2$ to $m$}
\STATE $\tilde{\textbf{H}}_{t-1}^* = \textbf{H}_{t}^*/(t-1)$
\STATE Initialize $W_t=I_k$, $i = 1$
  \WHILE{ not convergence}
  \STATE Fix $\textbf{B}$ and $\textbf{W}_t$, update $\textbf{E}$ by solving Eq.~\ref{optimizeE1}
  \STATE Fix $\textbf{E}$ and $\textbf{W}_t$, update \textbf{B} by solving Eq.~\ref{optimizeB1}
  \STATE Fix $\textbf{E}$ and $\textbf{B}$, update $\textbf{W}_t$ by solving Eq.~\ref{optimizeW1}
  \STATE $i=i+1$
  \ENDWHILE{$(obj^i-obj^{i-1})/obj^i$ $\leq$ $\varepsilon$}
  \STATE Update $\textbf{H}_t^*$ via Eq.~\ref{combine}
\ENDFOR
\STATE {\bfseries Output:} The final consensus partition matrix $\textbf{H}_m^*$.
   \smallskip
\end{algorithmic}
\end{algorithm}

\subsection{Three-step Alternate Optimization}
In the objective function Eq.~\ref{OBJ}, there are three terms that needs to be optimized, i.e., $\textbf{E}$, $\textbf{B}$ and $\textbf{W}_t$. To this end, we design a three-step alternate optimization to solve them.

\subsubsection{Optimizing $\textbf{E}$} Fixing $\textbf{B}$ and $\textbf{W}_t$, the optimization of Eq.~\ref{OBJ} can be formulated as follows,
\begin{equation}\label{optimizeE1}
%\begin{split}
 \max_{\textbf{E}} {\rm Tr} (\textbf{E}\textbf{A}^T) {\rm s.t.} \textbf{E}\in\{0,1\}^{n\times k}
%\end{split}
\end{equation}
where $ \textbf{A} = (\textbf{H}_t\textbf{W}_t + \lambda \tilde{\textbf{H}}_{t-1}^*)\textbf{B}^T$. Thus, the optimal $\textbf{E}$ can be represented as follows
\begin{equation}\label{optimizeE2}
%\begin{split}
 \textbf{E}(i,j) = 1
%\end{split}
\end{equation}
where $j = \arg \max A(i,:)$.

\subsubsection{Optimizing $\textbf{B}$} Fixing $\textbf{E}$ and $\textbf{W}_t$, the optimization of CAC objective function Eq.~\ref{OBJ} can be formulated as follows,
\begin{equation}\label{optimizeB1}
%\begin{split}
 \max_{\textbf{B}} {\rm Tr} (\textbf{B}^T\textbf{D}) {\rm s.t.} \textbf{B}^T\textbf{B} = \textbf{I}_k,
%\end{split}
\end{equation}
where $\textbf{D} = E^T(\textbf{H}_t\textbf{W}_t + \lambda \tilde{\textbf{H}}_{t-1}^*)$. The matrix $\textbf{D}$ can be formulated as $\textbf{D}=\textbf{S}\sum \textbf{V}^T$ by singular value decomposition (SVD), the optimization of $\textbf{B}$ can be rewritten as follows
\begin{equation}\label{optimizeB2}
%\begin{split}
\textbf{B} = \textbf{S} \textbf{V}^T
%\end{split}
\end{equation}
\subsubsection{Optimizing $\textbf{W}_t$} Fixing $\textbf{E}$ and $\textbf{B}$, the optimization of CAC objective function Eq.~\ref{OBJ} can be formulated as follows,
\begin{equation}\label{optimizeW1}
%\begin{split}
 \max_{\textbf{W}_t} {\rm Tr} (\textbf{W}_t^T\textbf{R})
%\end{split}
\end{equation}
where $\textbf{R}=\textbf{H}_t^T \textbf{E} \textbf{B}$. Similar with Eq.~\ref{optimizeB1}, Eq.~\ref{optimizeW1} can also be solved by SVD.

Now, we present the optimization process in Algorithm 1.

\subsection{Theoretical Analysis}
\subsubsection{Convergence}
The objective function Eq.~\ref{OBJ} can be separated into two terms as follows
\begin{equation}\label{eq2}
	\left\{
	\begin{aligned}
	    {\rm Tr}(\textbf{B}^T\textbf{E}^T \textbf{H}_t\textbf{W}_t) \\
        {\rm Tr}(\textbf{B}^T\textbf{E}^T\tilde{\textbf{H}}_{t-1}^*)
	\end{aligned}
	\right.
\end{equation}

By Cauchy-Schwartz inequality~\cite{BHATIA1995119}, we know ${\rm Tr}(\textbf{B}^T\textbf{E}^T \textbf{H}_t\textbf{W}_t)\leq ||\textbf{B}^T\textbf{E}^T||_F||(\textbf{H}_t\textbf{W}_t) ||_F = k$, where $k$ is the number of categories.

For the second term, $\textbf{B}^T\textbf{E}^T\tilde{\textbf{H}}_{t-1}^* \leq ||\textbf{B}^T\textbf{E}^T||_F ||\tilde{\textbf{H}}_{t-1}^*||_F$, since $||\tilde{\textbf{H}}_{t-1}^*||_F$ is a constant, we can obtain that $||\textbf{B}^T\textbf{E}^T||_F ||\tilde{\textbf{H}}_{t-1}^*||_F\leq a\sqrt{k}$ where $||\tilde{\textbf{H}}_{t-1}^*||_F =a$. Thus, we can obtain the upper bound of the CAC objective function Eq.~\ref{OBJ} as $k+a\sqrt{k}$. Besides, because our optimization method solves one variable while keeping other variables fixed, resulting in a monotonic increase in the value of objective function. Therefore, our algorithm is convergent. The experimental evidence also verifies the convergence of our algorithm.

\subsubsection{Complexity}
According to the optimization process in Algorithm, the time complexity in each iteration is $O(nk^2 + n)$, where $n$ is the number of data samples, $k$ is the number of clusters. There are $m$ views in the multi-view clustering setting, so the final time complexity of CAC algorithm can be seen as $O(lm(nk^2+n))$, where $l$ is the iteration number until convergence. For the space complexity, CAC algorithm only needs to store the category memory library $\textbf{B}$ and consensus partition matrix $\textbf{H}_t^*$. Thus, the space complexity is $O(nk+k^2)$.

\section{Experiments}

\subsection{Datasets}
\begin{figure}[t]
  \begin{center}
     \includegraphics[width=0.8\linewidth]{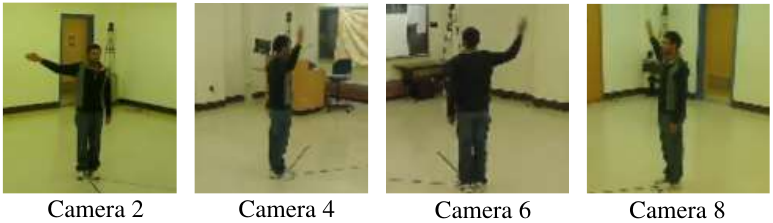}
  \end{center}
  \caption{Exemplar frames in WVU dataset}\label{data-example}
  %\vspace{-0.3cm}
\end{figure}

We adopt the following widely-used multi-view human action datasets to evaluate the effectiveness of the CAC. They are: \textbf{IXMAS}~\cite{6042901} consists of 11 different action categories, which has a total of 1, 695 video clips that recoded from 5 viewpoints. Due to the unavoidable partial occlusion, we select 4 views except for the top-down view. \textbf{MMI}~\cite{DBLP:journals/tcyb/LiuXNSWK17} consists of 22 interactive action categories with 1760 clips recorded in dark, light and cluttered environments. \textbf{MSR}~\cite{DBLP:conf/cvpr/XuMYR16} consists of 20 human action categories, each category is performed 2 or 3 times by 10 subjects. \textbf{WVU}~\cite{DBLP:conf/iccv/WeinlandBR07} contains 11 action categories and each category consists of 65 view clips. WVU is recorded from 8 cameras organized in a rectangular region. In this study, we select the actions that recorded non adjacent cameras, i.e., camera 2, 4, 6, 8, to make the incremental MVC more challenging. \textbf{UCLA}~\cite{DBLP:conf/cvpr/WangNXWZ14} consists of 10 daily actions recorded by 3 cameras. In total, it consists of 1475 RGB videos with depth and skeleton information. \textbf{MV-TJU}~\cite{DBLP:journals/ijon/LiuXSLHY15} consists of 22 human action categories and each category is performed 2 times by 20 subjects. Totally, it contains 3520 sequences with depth and skeleton modalities. Figure~\ref{data-example} shows exemplar frames in WVU dataset.

\subsection{Experimental Setup}
We implement the deep MVC baselines on PyTorch toolbox. Our proposed CAC and other multi-view clustering baselines are conducted on Matlab 2022a. All experiments are performed on a desktop computer with RTX 4090 GPU, i7-13700K CPU, 64G RAM, Windows 10 system.

\subsubsection{Baselines}
To fully demonstrate the effectiveness of the proposed CAC method, we compare it with the following baselines: late fusion MVC (LFMVC, IJCAI'19)~\cite{DBLP:conf/ijcai/WangLZTLHXY19}, one pass late fusion MVC (OPMVC, ICML'21)~\cite{DBLP:conf/icml/Liu0LWZTT0Z21}, fast multi-view clustering via ensembles (Fastmice, TKDE'23)~\cite{DBLP:journals/corr/abs-2203-11572}, multi-view subspace clustering via adaptive graph learning and late fusion alignment (AGLLFA, NN'23)~\cite{TANG2023333}, multi-view clustering with local kernel alignment (LKA, TMM'23)~\cite{DBLP:journals/tmm/ZhangLGWNS23}, simple multi-kernel $k$-means (SMKKM, TPAMI'23)~\cite{DBLP:journals/pami/Liu23}, binary multi-view clustering (BMVC, TPAMI'19)~\cite{8387526}, anchor-based partial multi-view clustering (APMC, AAAI'19)~\cite{DBLP:conf/aaai/GuoY19}, multi-kernel $k$-means clustering with matrix-induced regularization (Mir, AAAI'16)~\cite{DBLP:conf/aaai/LiuDY0Z16}, global and cross-view feature aggregation (GCFAgg, CVPR'23)~\cite{Yan_2023_CVPR}, adaptive feature projection with distribution alignment (APADC, TIP'23)~\cite{DBLP:journals/tip/XuLPRSSZ23}, auto-weighted orthogonal and nonnegative graph reconstruction (AONGR, INS'23)~\cite{DBLP:journals/isci/ZhaoYN23}, lifelong clustering (L2SC, TPAMI'22)~\cite{DBLP:journals/pami/SunCDLDY22}, incremental multi-view spectral clustering (SCGL, NN'21)~\cite{DBLP:journals/nn/YinHZLM21}, continual multi-view clustering (CMVC, MM'22)~\cite{DBLP:conf/mm/WanLLLWZ22}. Specifically, GCFAgg, APADC and AONGR are deep MVC baselines, L2SC, SCGL and CMVC are incremental MVC baselines and the remaining ones are MVC approaches based on shallow learning models.

\begin{table*}[]
\center
\footnotesize
\setlength{\tabcolsep}{0.85mm}
\begin{tabular}{c|ccc|ccc|ccc|ccc|ccc|ccc} \hline
         & \multicolumn{3}{c|}{IXMAS}       & \multicolumn{3}{c|}{MMI} & \multicolumn{3}{c|}{MSR} & \multicolumn{3}{c|}{WVU} & \multicolumn{3}{c|}{UCLA} & \multicolumn{3}{c}{MVTJU} \\ \cline{2-19}
         & ACC   & NMI            & Purity & ACC    & NMI   & Purity & ACC    & NMI   & Purity & ACC    & NMI   & Purity & ACC    & NMI    & Purity & ACC    & NMI    & Purity  \\ \hline
LFMVC    & 65.76 & \textbf{66.56} & 65.76  & 43.44  & 53.97 & 44.61  & 41.88  & 48.21 & 45.31  & 65.69  & 62.42 & 65.69  & 32.29  & 37.54  & 36.72  & 68.18  & 79.63  & 71.16   \\
OPMVC    & 62.73 & 64.65          & 64.85  & 39.23  & 49.80  & 42.15  & 40.63  & 47.46 & 45.31  & 49.85  & 50.14 & 52.46  & 40.89  & 42.30   & 43.75  & 64.89  & 75.78  & 65.82   \\
Fastmice & 56.67 & 61.05          & 59.57  & 38.19  & 49.8  & 40.42  & 40.62  & 48.11 & 45.18  & 58.46  & 62.10  & 60.58  & 37.13  & 43.55  & 42.44  & 64.43  & 77.68  & 66.57   \\
AGLLFA   & 60.30  & 64.73          & 61.82  & 45.67  & 56.45 & 48.36  & 39.69  & 41.49 & 40.31  & 65.23  & 66.76 & 66.77  & 42.45  & 44.80   & 47.92  & 65.51  & 75.86  & 67.67   \\
LKA      & 65.45 & 64.49          & 65.45  & 45.20   & 54.40  & 46.72  & 46.56  & 50.23 & 47.81  & 68.92  & 65.52 & 69.08  & 44.01  & 43.87  & 45.31  & 71.56  & 79.16  & 73.10    \\
SMKKM    & 63.56 & 65.21          & 64.38  & 44.60   & 55.46 & 46.86  & 41.72  & 47.59 & 44.62  & 65.18  & 61.33 & 65.22  & 32.06  & 36.53  & 36.82  & 68.47  & 79.46  & 70.58   \\
BMVC     & 57.64 & 60.32          & 63.12  & 44.26  & 55.14 & 47.12  & 40.32  & 44.36 & 44.23  & 56.12  & 53.12 & 65.12  & 30.26  & 31.26  & 34.26  & 68.45  & 78.56  & 70.78   \\
APMC     & 58.79 & 61.50           & 65.76  & 46.60   & 56.41 & 48.12  & 42.19  & 45.97 & 43.44  & 55.69  & 56.51 & 64.00   & 29.95  & 31.56  & 34.11  & 70.94  & 79.68  & \textbf{75.54}   \\
MIR      & 62.12 & 63.64          & 62.73  & 43.44  & 53.97 & 44.61  & 42.81  & 46.62 & 46.56  & 65.85  & 62.59 & 65.54  & 32.55  & 37.87  & 36.98  & 69.80   & 79.46  & 70.57   \\
GCFAgg   & 58.48 & 60.61          & 59.09  & 39.23  & 51.47 & 42.51  & 37.19  & 42.00    & 38.12  & 60.31  & 66.78 & 63.85  & \textbf{44.95}  & 43.73  & 47.03  & 67.87  & 77.34  & 67.98   \\
APADC    & 55.63 & 58.36          & 57.45  & 40.78  & 52.15 & 43.21  & 40.36  & 44.36 & 40.19  & 63.15  & 61.36 & 59.36  & 43.69  & 44.43  & 43.26  & 63.54  & 71.35  & 67.46   \\
AONGR    & 43.96 & 49.29          & 45.15  & 29.31  & 40.60  & 30.98  & 39.06  & 46.80  & 42.80   & 39.38  & 55.74 & 44.92  & 38.02  & 44.46  & 44.27  & 63.40   & 77.67  & 66.70    \\
L2SC     & 61.67 & 62.11          & 62.34  & 44.67  & 55.24 & 47.42  & 41.72  & 46.77 & 44.06  & 64.46  & 64.64 & 65.42  & 27.34  & 23.36  & 29.94  & 69.87  & 79.06  & 71.12   \\
SCGL     & 61.52 & 61.85          & 61.82  & 42.27  & 54.30  & 43.21  & 40.63  & 45.29 & 44.06  & 66.00     & 66.60  & 66.77  & 36.20   & 37.88  & 40.36  & 68.21  & 79.30   & 69.72   \\
CMVC     & 63.33 & 63.58          & 63.94  & 44.50   & 54.77 & 46.84  & 46.88  & 51.83 & 49.06  & 69.69  & 65.17 & 69.69  & 42.71  & 44.32  & 47.14  & 71.99  & 79.94  & 72.87   \\ \hline
CAC      & \textbf{68.79} & 65.77          & \textbf{68.79} & \textbf{46.96} & \textbf{56.63} & \textbf{48.48} & \textbf{49.06} & \textbf{52.30} & \textbf{51.25} & \textbf{73.85} & \textbf{70.53} & \textbf{73.85} & 44.79 & \textbf{48.52} & \textbf{48.44} & \textbf{73.69} & \textbf{80.30} & 74.18 \\ \hline
\end{tabular}
\caption{Comparisons on the 6 datasets in terms of ACC, NMI and Purity (mean results). Bold font denotes best values.} \label{ComparisonResults}
\end{table*}

\subsubsection{Parameter Setting}
In our model, there is a regularization parameter $\lambda$ that balances the weights between historical and new coming views. Similar with the late fusion MVC baselines (CMVC, LFMVC, LKA), we tune $\lambda$ in the range of $2.^{\wedge}[-10,-9,\cdots,9, 10]$. We perform all baselines according to the original parameter settings from their papers, and the source codes are available from original authors or websites.
We adopt ACC, NMI and Purity to measure the clustering performance as in~\cite{8387526,DBLP:journals/pami/Liu23,DBLP:journals/tmm/ZhangLGWNS23}. Higher value indicates better clustering performance. For fairness, we run all methods 10 times and report the average results.

\subsection{Experimental Results and Analysis}
Table~\ref{ComparisonResults} shows the comparison results of CAC and baselines on 6 human action data collections. We can obtain the following observation from these tables.

%\begin{table*}[]
%\caption{Comparisons on the 6 datasets in terms of Purity \% (mean results). Best results are bolded.} \label{Purity}
%\scriptsize
%\centering
%\setlength{\tabcolsep}{1.45mm}
%\begin{tabular}{ccccccccccccccccc}
%\hline
%      & LFMVC & OPMVC & Fastmice & AGLLFA & LKA   & SMKKM & BMVC  & APMC  & MIR & GCFAgg & APADC & AONGR & L2SC  & SCGL  & CMVC  & CAC   \\
%      \hline
%IXMAS & 65.76 & 64.85 & 59.57    & 61.82  & 65.45 & 64.38 & 63.12 & 65.76 & 62.73 & 59.09  & 57.45 & 45.15  & 62.34 & 61.82 & 63.94 & \textbf{68.79} \\
%MMI   & 44.61 & 42.15 & 40.42    & 48.36  & 46.72 & 46.86 & 47.12 & 50.12 & 44.61 & 42.51  & 43.21 & 30.98 & 47.42 & 43.21 & 46.84 & \textbf{48.48} \\
%MSR   & 45.31 & 45.31 & 45.18    & 40.31  & 47.81 & 44.62 & 44.23 & 43.44 & 46.56 & 38.12  & 40.19 & 42.80 & 44.06 & 44.06 & 49.06 & \textbf{51.25} \\
%WVU   & 65.69 & 52.46 & 60.58    & 66.77  & 69.08 & 65.22 & 65.12 & 64.00 & 65.54 & 63.85  & 59.36 & 44.92 & 65.42 & 66.77 & 69.69 & \textbf{73.85} \\
%UCLA  & 36.72 & 43.75 & 42.44    & 47.92  & 45.31 & 36.82 & 34.26 & 34.11 & 36.98 & 47.03  & 43.26 & 44.27 & 29.94 & 40.36 & 47.14 & \textbf{48.44} \\
%MVTJU & 71.16 & 65.82 & 66.57    & 67.67  & 73.10 & 70.58 & 70.78 & 75.54 & 70.57 & 67.98  & 67.46 & 66.70  & 71.12 & 69.72 & 72.87 & \textbf{74.18} \\
%\hline
%\end{tabular}
%%\end{center}
%\end{table*}

1) The proposed CAC outperforms the traditional MVC baselines. For example, as an extension of late fusion MVC (LFMVC), the proposed CAC obtains 3.03\%, 3.52\%, 7.18\%, 8.16\%, 12.5\% and 5.51\% improvements respectively on the 6 datasets used in this paper. This is mainly because the proposed CAC well capture the consistency between multiple views in a continual clustering manner.

2) Recently, utilizing DNNs to boost the performance of MVC has received attractive attentions. To further verify the effectiveness of the CAC, we compare it with several latest state-of-the-art deep MVC baselines. From Table~\ref{ComparisonResults}, we can observe that the CAC also performs better than the deep MVC baselines. For example, the CAC obtains 5.16\%, 7.41\%, 16.48\% improvements compared with GCFAgg, APADC and AONGR  on IXMAS dataset in terms of NMI. This phenomenon further verifies the effectiveness of the CAC.

3) Compared with other incremental MVC baselines (L2SC, SCGL and CMVC), the CAC also achieves better performance or even a significant improvement on the 6 human action datasets. For example, CAC obtains 7.12\%, 7.27\% and 5.46\% improvements compared with L2SC, SCGL and CMVC respectively on the IXMAS dataset in term of ACC metric. This is mainly because the knowledge of historical views is easily forgotten in the SCGL and CMVC, since they do not have a knowledge library to dynamically update the knowledge of historical views. L2SC requires to maintain two banks, i.e., feature bank and clustering partition bank, which is time-consuming making (see the subsection of running time analysis) and it is impratical for human action clustering applications. In this study, we design a category memory library. When new action view is coming, we only need to maintain a consensus partition matrix, which can be updated by leveraging the incoming new camera view rather than keeping all of them.
\begin{figure}[t]
  \begin{center}
     \includegraphics[width=1.0\linewidth]{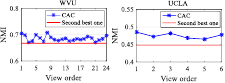}
  \end{center}
  \caption{Impact of view order on CAC.}\label{view-order}
  %\vspace{-0.3cm}
\end{figure}

\begin{figure}[t]
  \begin{center}
     \includegraphics[width=0.9\linewidth]{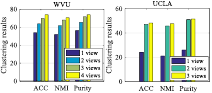}
  \end{center}
  \caption{Impact of view number on CAC.}\label{viewnum}
  %\vspace{-0.3cm}
\end{figure}
\subsection{Impact of View Order}
In the scenario of incremental MVC, the views are incremental over time. Thus, it is necessary to investigate the impact of the order of view arrival on the clustering performance of the CAC. For example, given a multi-view data collection with $m$ views, there are $m! = m*(m-1)*\cdots*1$ number of orders that integrates the views. In this subsection, we report the clustering results of CAC with different view orders on WVU and UCLA datasets. As shown in Figure~\ref{view-order}, the CAC outperforms consistently the second best baselines on these two datasets. We can conclude that the view order has little impact on the final clustering performance.%, which also verifies the robustness of CAC on the different view orders.
\begin{figure}[t]
  \begin{center}
     \includegraphics[width=1.0\linewidth]{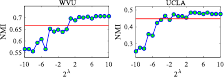}
  \end{center}
  \caption{Various parameter $\lambda$ on CAC.}\label{para}
  %\vspace{-0.3cm}
\end{figure}

\begin{figure}[t]
  \begin{center}
     \includegraphics[width=1.0\linewidth]{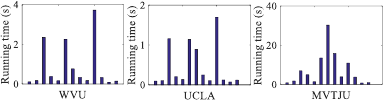}
  \end{center}
  \caption{Running time comparison of our CAC with other baselines. In each sub-figure, the baselines in x-axis are LFMVC, OPMVC, Fastmice, AGLLFA, LKA, SSMKKM, BMVC, APMC, MIR, L2SC, SCGL, CMVC and our CAC.}\label{time}
  %\vspace{-0.3cm}
\end{figure}
\subsection{Impact of View Number}
Since the CAC is an incremental MVC approach, which is capable of dealing with views over time. Ideally, as a new view joins, the clustering performance of incremental multi-view clustering method should increase rather than decrease. In this subsection, we investigate the impact of view number on the clustering performance of CAC. We only report the experimental results on WVU and UCLA datasets due to page limit. From Figure~\ref{viewnum}, we can see that the performance of CAC unintentionally improves as the view number increases. This phenomenon verifies that the CAC can well leverage the knowledge of historical views to assist the clustering of new coming views.

\begin{figure}[t]
    \centering
    \subfigure[WVU]{
        \includegraphics[width=0.20\textwidth]{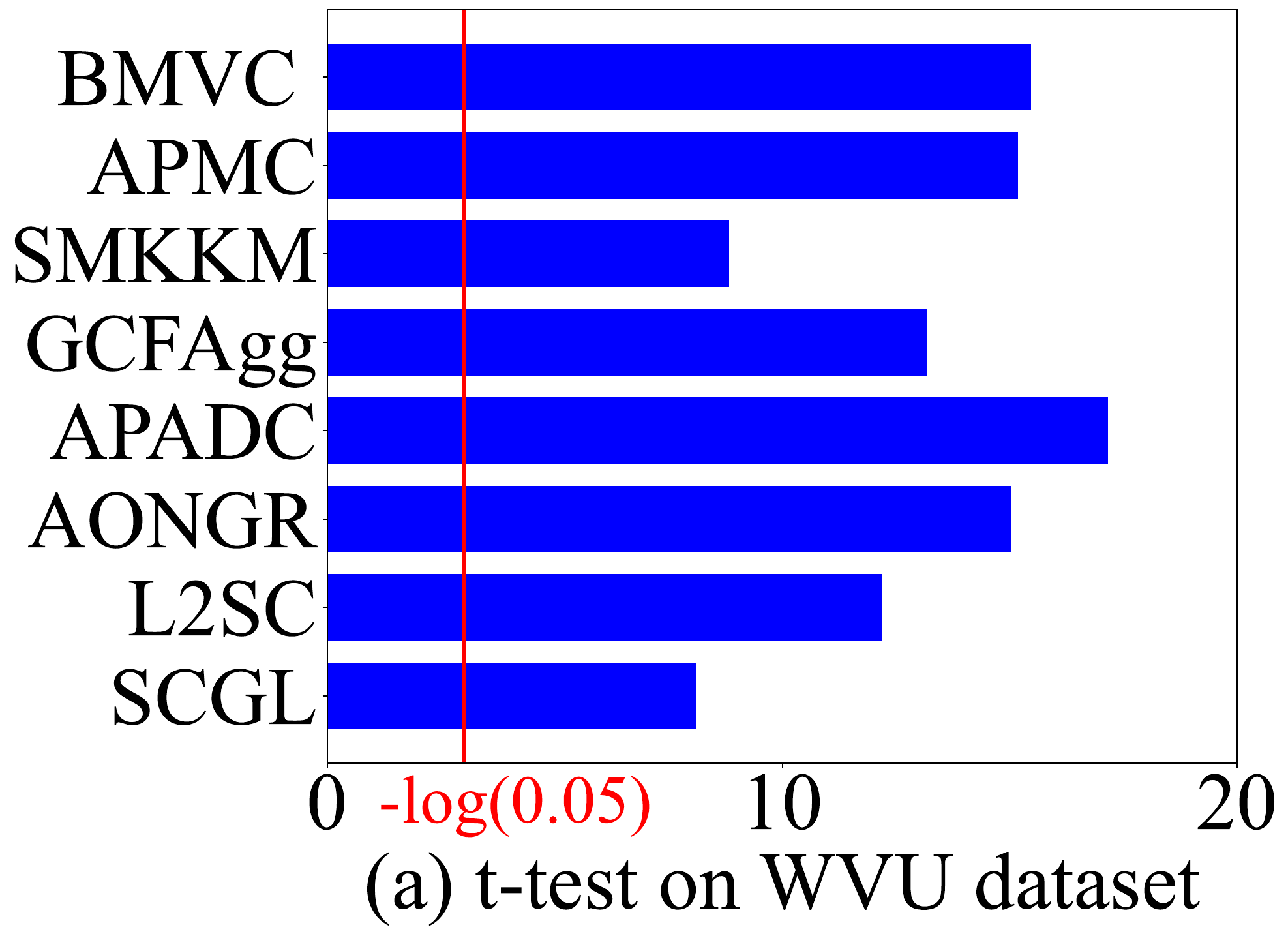}
    }
    \subfigure[UCLA]{
        \includegraphics[width=0.20\textwidth]{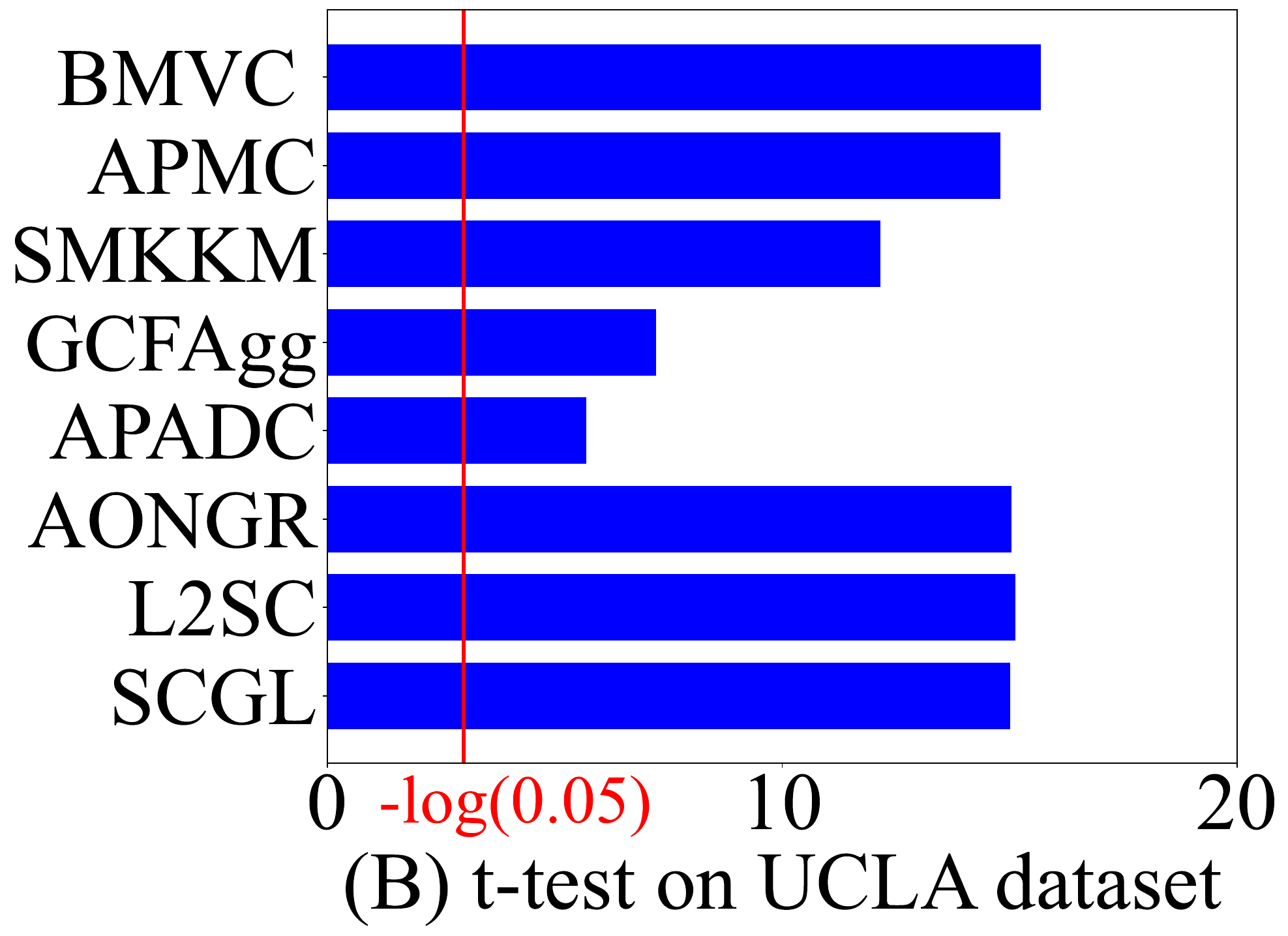}
    }
    \caption{Significance $t$-test on WVU and UCLA (ACC).}
    \label{significance}
\end{figure}
\subsection{Parameter Investigation}
In this subsection, we conduct experiments to investigate the impact of the regularization parameter $\lambda$ on the clustering performance of CAC method. Specifically, we vary the value of $\lambda$ in the range of $2.^{\wedge}[-10,-9,\cdots,9, 10]$. We only report the experimental results on WVU and UCLA due to page limit. Figure~\ref{para} provides the comparisons of CAC with the second best baseline. From this figure, we can observe that CAC trends to be better and stable when $\lambda$ takes large value in the range. This experiment demonstrates that the CAC is stable and does not depend on precise parameter tuning.

\subsection{Running Time Analysis}
In this subsection, we investigate the running efficiency of the CAC method. For fairness, we only compare the running time of CAC with the baselines based on shallow learning models since deep MVC methods usually take too much time. As shown in Figure~\ref{time}, although the CAC is not the fastest one, its running time is comparable with the most efficient ones. This verifies that the CAC takes much less time to deal with the multi-view data.

\subsection{Significance Study}
We carry out $t$-test study on WVU and UCLA with several state-of-the-art baselines in terms of ACC, in which the $p$-value is processed by $-\log(p)$ with a significance level of 0.05. From Figure~\ref{significance}, we observe that the CAC is statistically significant compared with these baselines.

\begin{figure}[t]
  \begin{center}
     \includegraphics[width=1.0\linewidth]{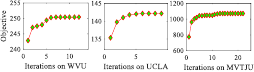}
  \end{center}
  \caption{Convergence curves of the CAC on the task of incremental multi-view clustering.}\label{convergence}
  %\vspace{-0.3cm}
\end{figure}

\subsection{Convergence}
As shown in Figure~\ref{convergence}, with the increase in experimental iterations, the objective function of the CAC monotonically approaches a fixed point on WVU, UCLA and MVTJU datasets. This phenomenon demonstrates that the CAC converges quickly and stably.% in the task of incremental multi-view action clustering.

\subsection{Adaptation to other MVC Datasets}
\begin{figure}[t]
  \begin{center}
     \includegraphics[width=1.0\linewidth]{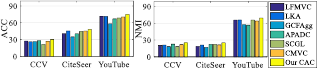}
  \end{center}
  \caption{Comparison with representative traditional, deep and incremental MVC baselines on 3 MVC datasets.}\label{easy}
\end{figure}

To facilitate easier comparison for readers, we apply the CAC to more datasets widely used in the MVC baselines, i.e., CCV~\cite{DBLP:conf/mir/JiangYCEL11}, CiteSeer\footnote{https://linqs-data.soe.ucsc.edu/public/lbc/citeseer.tgz} and YouTube~\cite{DBLP:journals/ml/MadaniGR13}. Figure~\ref{easy} shows the comparison results with several traditional, deep and incremental MVC baselines, which further verifies the effectiveness and robustness of the CAC.

\section{Conclusions}
This paper investigates a continual action clustering task with incremental views and proposes a novel CAC method, which achieves never-ending knowledge transfer between historical views and the new coming ones by leveraging late fusion MVC, and improves the clustering performance accordingly. Extensive experimental results demonstrate the superior performance and time/space efficiency of the CAC compared with 15 state-of-the-art baselines. %In future, it is insightful to explore more promising and challenging incremental or life-long learning tasks with CAC.

\section{Acknowledgments}
This work is supported by the National Natural Science Foundation of China under grant No. 61906172, China Postdoctoral Science Foundation under grant No. 2020M682357 and Key Science and Technology Program of Henan Province under grant No. 222102210012.
\bibliography{aaai24}

\end{document}